\documentclass[journal]{IEEEtran}

\ifCLASSINFOpdf
\else
\fi

\usepackage{float}
\usepackage{amssymb}
\usepackage{amsmath}
\usepackage{hyperref}
\usepackage{url}
\usepackage{array}
\usepackage{multirow}
\usepackage[table,xcdraw]{xcolor}
\usepackage{multirow}
\usepackage{graphicx}

\begin{document}

\title{SegStitch: Multidimensional Transformer for Robust and Efficient Medical Imaging Segmentation}
\author{Shengbo Tan,
        Zeyu Zhang,
        Ying Cai,
        Daji Ergu,
        Lin Wu,
        Binbin Hu,
        Pengzhang Yu,
        Yang Zhao
\thanks{Ying Cai is the corresponding author: \url{caiying34@yeah.net}.}
\thanks{Shengbo Tan, Ying Cai, Daji Ergu, Binbin Hu, and Pengzhang Yu are with Key Laboratory of Electronic Information Engineering, College of Electronic and Information, Southwest Minzu University, Chengdu 610225, China.}
\thanks{Zeyu Zhang is with College of Engineering, Computing and Cybernetics, the Australian National University, Canberra ACT 2601, Australia.}
\thanks{Lin Wu is with Hunan University of Technology, Zhuzhou 410072, China.}
\thanks{Yang Zhao is with Department of Computer Science and Information Technology, La Trobe University, Bundoora VIC 3086, Australia.}}

\markboth{Journal of \LaTeX\ Class Files,~Vol.~14, No.~8, August~2015}%
{Shell \MakeLowercase{\textit{et al.}}: Bare Demo of IEEEtran.cls for IEEE Journals}

\maketitle

\newcommand{\qq}[1]{\textcolor{blue}{[Zeyu: #1]}}

\begin{abstract}
Medical imaging segmentation plays a significant role in the automatic recognition and analysis of lesions.  State-of-the-art methods, particularly those utilizing transformers, have been prominently adopted in 3D semantic segmentation due to their superior performance in scalability and generalizability. However, plain vision transformers encounter challenges due to their neglect of local features and their high computational complexity. To address these challenges, we introduce three key contributions: Firstly, we proposed SegStitch, an innovative architecture that integrates transformers with denoising ODE blocks. Instead of taking whole 3D volumes as inputs, we adapt axial patches and customize patch-wise queries to ensure semantic consistency. Additionally, we conducted extensive experiments on the BTCV and ACDC datasets, achieving improvements up to 11.48\% and 6.71\% respectively in mDSC, compared to state-of-the-art methods. Lastly, our proposed method demonstrates outstanding efficiency, reducing the number of parameters by 36.7\% and the number of FLOPS by 10.7\% compared to UNETR. This advancement holds promising potential for adapting our method to real-world clinical practice. The code will be available at \href{https://github.com/goblin327/SegStitch}{https://github.com/goblin327/SegStitch}

\end{abstract}


\IEEEpeerreviewmaketitle

\section{Introduction}

3D segmentation stands as a pivotal challenge within the medical imaging domain, with broad implications for clinical practices. Notably, delineating the intricate internal structures of the human body, including organs, within medical imagery is essential for key clinical applications like Computer-Aided Diagnosis (CAD), Computer-Assisted Surgery (CAS), and Radiotherapy (RT) \cite{56}.

In recent years, the mainstream approaches to medical image segmentation have predominantly relied on Convolutional Neural Networks (CNN) \cite{1}, such as U-Net \cite{2} and its related architectures \cite{3,4,5,6,7}. 
However, these methods often struggle with convolutional localization, limiting their ability to effectively capture contextual information\cite{58}.To address this limitation, there is an urgent need for efficient medical segmentation networks. However, ViT \cite{8} has demonstrated state-of-the-art performance in various computer vision \cite{ji2024sine,wu2024xlip} and medical image analysis tasks \cite{zhang2024jointvit}. 
Firstly, ViT \cite{9} leverages its outstanding capability in capturing long-range interactions, demonstrating excellent generalization and robust features \cite{10}. Secondly, this approach exhibits high parallelism, facilitating the training and inference processes of large-scale models \cite{12}.

In terms of algorithm improvements, a focal point of attention has been the enhancement of core building blocks, particularly focusing on attention mechanisms. However, such performance comes at a cost, due to attention requiring the computation of feature information across all spatial positions, it incurs high computational complexity and significant memory overhead. Nevertheless, queries in different semantic regions actually involve entirely different key-value pairs, so forcing all queries to attend to the same token set may be suboptimal. 
Therefore, we extended the self-attention module by dividing the attention mechanism into fine-grained and coarse-grained modules, and enabling them to work in tandem. This approach further enhances the model's ability to understand and represent complex features.

\begin{figure}
\centering
\includegraphics[width=1\linewidth]{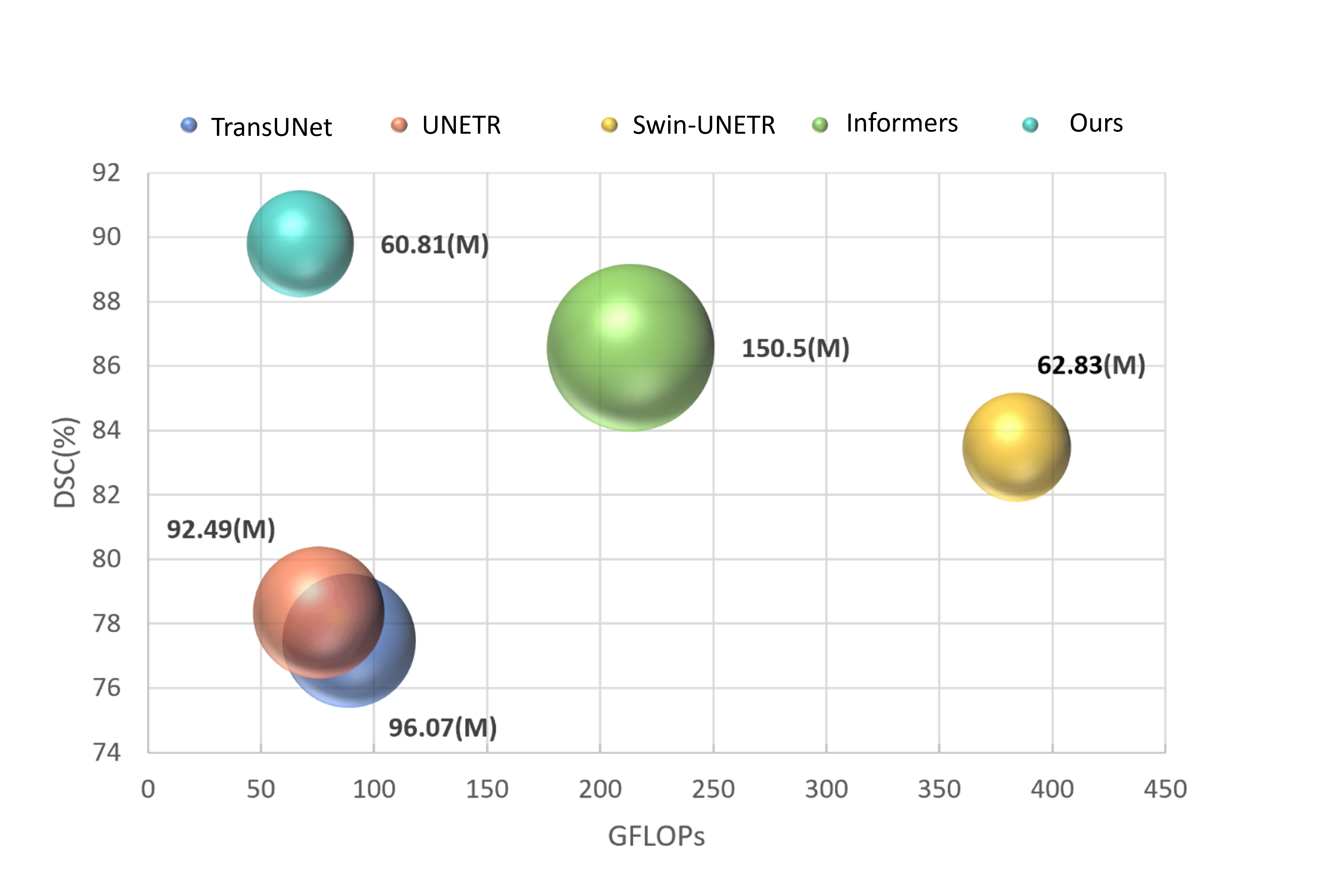}
\caption{\label{fig:flop}Relationship between model parameter count, computational complexity, and Dice similarity coefficient. The size of the spheres indicates the model parameter count. Compared to other models, our SegStitch achieves the highest mDSC while maintaining smaller model size and lower computational complexity.}
\end{figure}

According to research findings, in neural ODE-based networks (ODENets), even for data subjected to slight perturbations, the difference between the corresponding perturbed version of the integral curve and the original curve is not significant \cite{18}. 
Inspired by Neural Memory Ordinary Differential Equations (nmODE) \cite{35}, we have integrated the SegStitch module with nmODE to facilitate the dynamic and continuous updating of feature maps, leading to further optimized segmentation results.

To further assess our proposed method, We validate our SegStitch method using two datasets: Synapse \cite{19} and ACDC \cite{20}. Both qualitative and quantitative results demonstrate the effectiveness and superior segmentation capability of SegStitch in terms of segmentation accuracy and model efficiency. On the Synapse dataset, SegStitch achieves high-quality segmentation of long-range targets (see Figure \ref{fig:show7}), with an absolute gain of 0.97\% in Dice Score (as shown in Table \ref{tab:Synapse}), while significantly reducing model complexity. Compared to the baseline UNETR, SegStitch reduces parameters by 36.7\% and FLOPs by 10.7\% \cite{21},As shown in Figure \ref{fig:flop}.

Our contributions can be summarized as follows:

(1) First, we propose \textbf{SegStitch}, an innovative architecture that combines transformers with denoising ODE blocks. In this transformer, we customized shared queries between different patches to enhance contextual relationships using shared query keys, ensuring semantic consistency. This achieves high-quality segmentation of long-range targets through local-to-global self-attention.

(2) Additionally, we conducted extensive experiments on the BTCV and ACDC datasets. Compared to state-of-the-art methods, we achieved improvements of \textbf{11.48\%} and \textbf{6.71\%} in the mDSC metric, respectively.

(3) Finally, our proposed method demonstrates outstanding efficiency. Compared to UNETR, it achieves a \textbf{36.7\%} reduction in the number of parameters and a \textbf{10.7\%} reduction in FLOPS. These advancements suggest promising prospects for applying our method in practical clinical settings.

\section{Related Work}

\subsection{3D Medical Image Segmentation}

Three-dimensional medical image segmentation is a crucial technique that distinguishes organs and pathological areas from medical images on a pixel-by-pixel basis. The development of this technology has greatly benefited from the remarkable achievements of Convolutional Neural Networks (CNNs) \cite{1}. Typical work, such as 3D U-Net and nnU-Net \cite{isensee2021nnu}, has been widely adopted for medical imaging segmentation tasks \cite{wu2023bhsd,zhangthin,zhang2023segreg}. Researchers have been actively exploring the integration of Transformers alongside differential neural networks to overcome the limitations of CNNs and enhance segmentation accuracy. In recent years, significant efforts have been made to integrate these technologies into medical image segmentation, aiming to maximize accuracy. In the following sections, we will introduce and analyze the advancements made in the domain of three-dimensional medical image segmentation through the combination of CNNs, Transformers, and differential neural networks.

To address the limitation of local receptive fields in convolutional neural networks, researchers have explored the use of dilated convolutions and large kernel convolutions as alternatives to standard convolutions \cite{59}. Inspired by the significant success of Transformers in natural language processing, methods like TransUNET \cite{60} and similar approaches \cite{61} \cite{62} \cite{63}  combine Transformers with convolutions as building blocks for encoders. Transclaw Unet \cite{45} integrates convolutional operations with Transformers on the encoder side, achieving detailed segmentation and long-range relationship learning. Meanwhile, Azad et al. \cite{64} introduced TransDeepLab, which enhances DeepLab with different window strategies for skin lesion segmentation. They also introduced TransCeption \cite{65}, which captures multi-scale representations in a single stage through an improved patch merging module. Additionally, Huang et al. \cite{66} enhanced segmentation performance by integrating global information across different scales within a U-shaped architecture.Wu et al. \cite{23} achieved more efficient feature extraction and representation through patch embedding processing. Ali et al. \cite{25} introduced an encoder based on 3D Swin Transformers and a decoder that combines CNN and Transformer-based techniques, effectively reconstructing high-quality segmentation images. Yun et al. \cite{26} proposed an enhanced Transformer module to connect the encoder and decoder, effectively reducing information loss between hierarchical features. Li et al. \cite{28} proposed an innovative upsampling method that simulates long-range dependencies and integrates global information, capturing and preserving spatial edge details more accurately. Shaker et al. \cite{29} introduced the Efficient Pairwise Attention (EPA) module, which effectively learns features through spatial and channel-based attention.

Currently, methods based on Vision Transformers (ViT) have achieved significant success in medical image segmentation, primarily due to their strong representation capabilities. However, Transformers face challenges in handling long-range dependencies, as the attention mechanism requires calculating attention between all positions, leading to increased computational complexity with longer sequences. Furthermore, the sparsity of medical image data makes models prone to overfitting during training. An underexplored area is the application of multimodal attention mechanisms to reveal long-range edge information in three-dimensional medical images. Due to the inherent complexity of medical images, extracting long-range edge information remains a relatively unexplored and challenging topic.

\subsection{Ordinary Differential Neural Network}

In recent years, only a few segmentation methods have combined neural ODEs with network design \cite{67} \cite{68} \cite{69}. In a study \cite{30}, the structure of U-Net was enhanced by introducing elements of neural ODEs. In this design, the repetitive residual blocks in each branch of the U-Net were replaced by neural ODEs, which operated only around a convolutional block. Despite the widespread recognition of U-Net, recent research indicates that new network designs can achieve superior results compared to U-Net. A recent study \cite{31} proposed an innovative approach that combines neural ODEs with the level set method. This method utilizes parameterized derivative contours as neural ODEs, implicitly learning the governing functions describing contour evolution. Additionally, several studies \cite{32},\cite{33},\cite{34} have applied neural ODEs to defense against adversarial attacks, achieving notably significant results.

Building upon existing research, our work incorporates the use of neural memory ordinary differential equation blocks into the last fully connected layer of the Vision Transformer (VIT) architecture. This innovative design has achieved state-of-the-art performance on multiple datasets, demonstrating superior stability and reliability compared to traditional Convolutional Neural Networks (CNNs). By integrating neural memory ordinary differential equation with VIT, we have achieved significant improvements in medical segmentation tasks, providing a new paradigm for the application of neural memory ordinary differential equation in network design.

\begin{figure}
\centering
\includegraphics[width=1\linewidth]{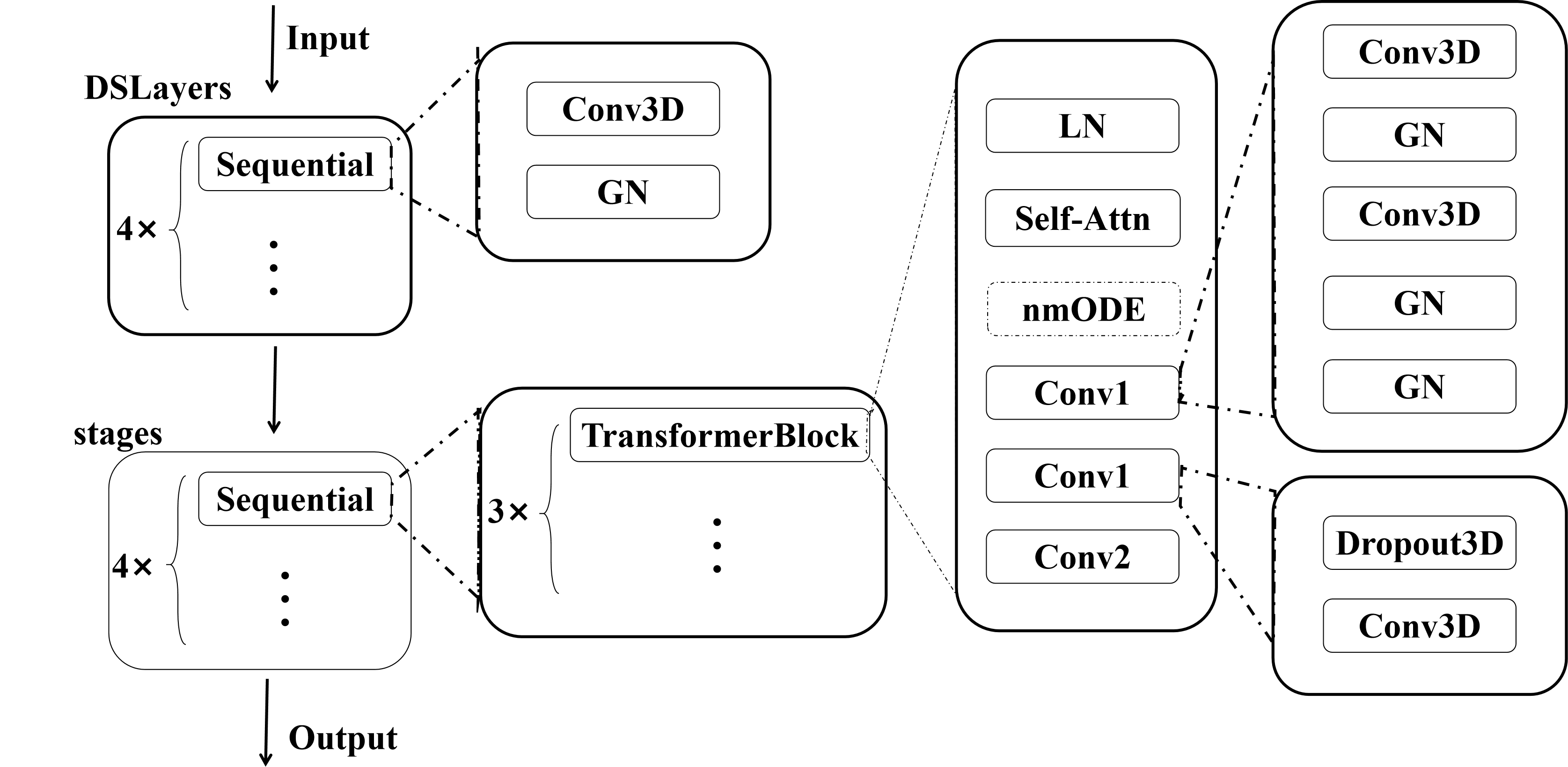}
\caption{\label{fig:DSL}SegStitch Network Encoder Structure Diagram. The entire network comprises four layers, each containing a downsampling module and stages. The downsampling module consists of convolution and Group Normalization, while the stages consist of three Transformer Blocks. Each Transformer Block computes different patches, together forming the complete downsampling module.}
\end{figure}

\begin{figure}
\centering
\includegraphics[width=0.7\linewidth]{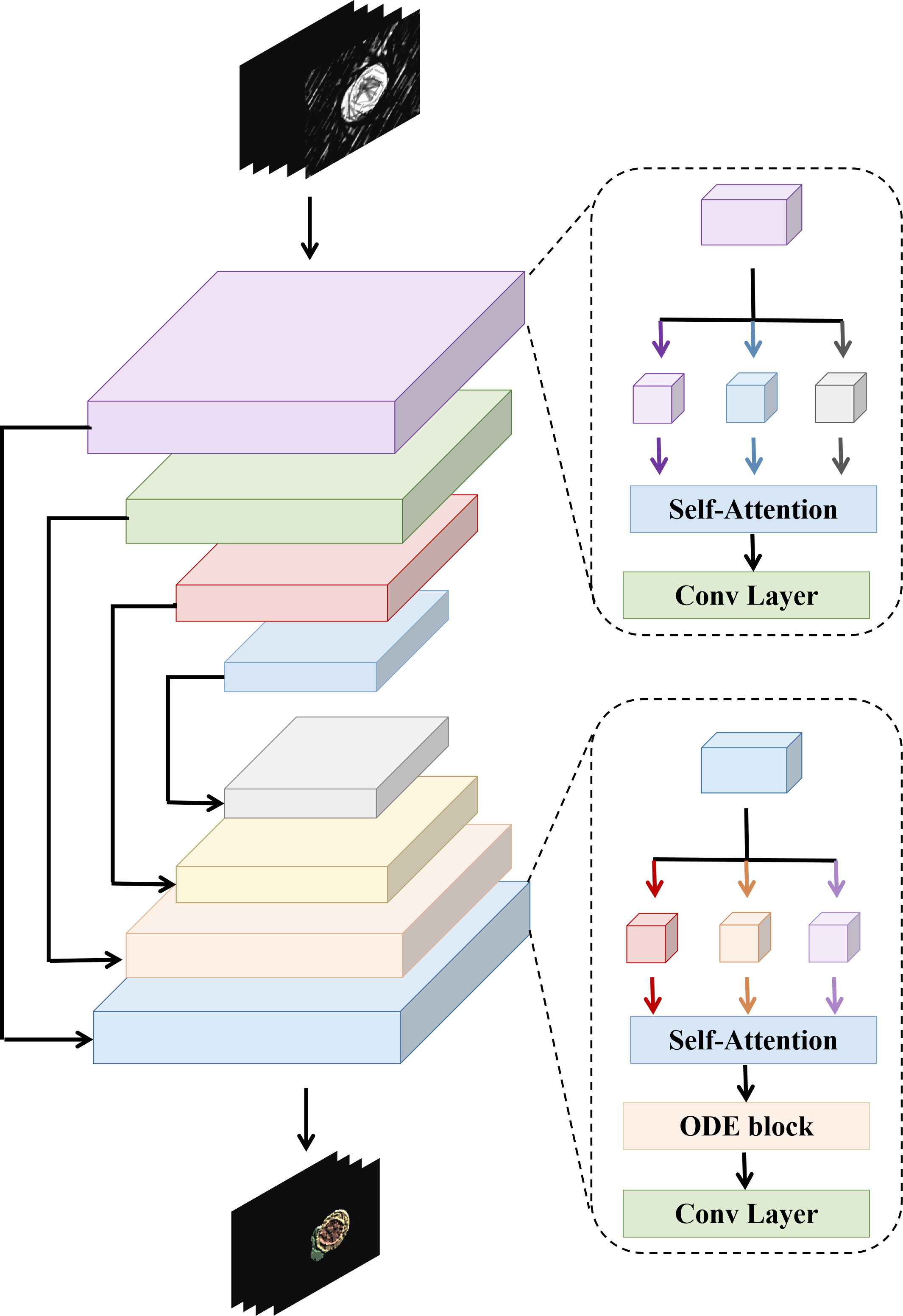}
\caption{\label{fig:main_1}Overall Architecture. The SegStitch method utilizes a hierarchical encoder-decoder structure. The output of the downsampling modules is passed to the decoder through skip connections, and each decoder module generates the final segmentation mask using ODE blocks.}
\end{figure}

\section{Method}

\subsection{Overall Architecture}

We propose SegStitch, a novel 3D image segmentation network. The overall architecture of SegStitch is illustrated in Figure \ref{fig:main_1}. Our approach enhances traditional methods by improving the positional information weighting of each pixel in the 3D image. We then introduce a dedicated 3D image feature extraction module (as shown in Figure \ref{fig:DSL}) to effectively extract features from 3D data. Finally, the outputs of the self-attention mechanism are fed into a Neural Memory Ordinary Differential Equation network, aimed at improving the model's generalization capability and stability.Specifically, given a multi-modal MRI medical image input $X \in \mathbb{R}^{M \times C \times H \times W \times D}$ (where (H, W, D) denotes the input resolution, M represents the number of modalities, and C denotes the number of channels), the encoder layers process the input starting from the first layer. The output of each encoder layer is processed through the SegStitch module, generating planar feature maps with half the dimensions of the input. Consequently, at the final layer, we obtain a global feature vector of size [1×64×156]. The decoder layers then convert the dense features back into segmentation image planes, with upsampled features concatenated with the outputs from the corresponding encoder layers to better distinguish between foreground and background pixels. This results in a more robust segmentation mask.A key aspect of our design is the introduction of the effective feature extraction block, SegStitch. In the SegStitch module, self-attention employs parallel attention modules to generate diverse features through shared queries, thus effectively learning rich feature representations. Finally, the outputs from the two attention modules are fused and then fed into the ODE block to generate high-quality 3D medical image masks.

\begin{figure}
\centering
\includegraphics[width=0.8\linewidth]{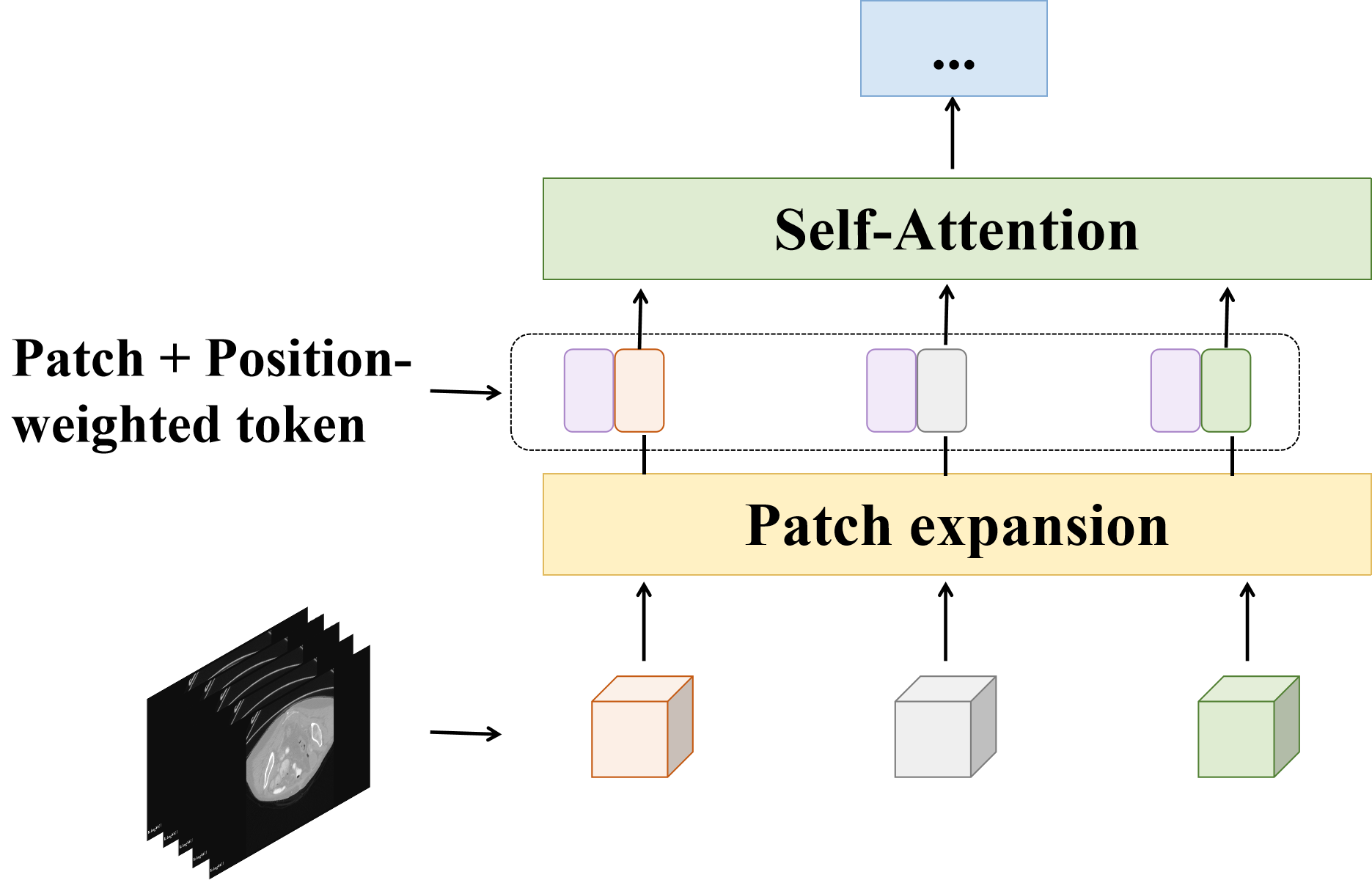}
\caption{\label{fig:token}Position-weighted token. we generate a set of learnable parameters $\text{Token}_{wt}$ with a size of 1×d (where d represents the total length after flattening the 3D patch). These parameters are used to learn the spatial positional information of the 3D patch.}
\end{figure}
\subsection{Position-Weighted Tokens}

To address the loss of spatial positional information when directly flattening 3D patches before input tokenization, we generate a set of learnable parameters $\text{Token}_{wt}$ with a size of 1×d (where d represents the total length after flattening the 3D patch). we sum them with $\text{Token}_{org}$ to obtain $\text{Token}_{final}$, which possesses positional information (as illustrated in Figure \ref{fig:token}). The definition of $\text{Token}_{final}$ is as follows:

\begin{equation}
\text{Token}_{final} = \text{ADD}(\text{Token}_{org}, \text{Token}_{wt})
\end{equation}

Where $\text{Token}_{final}$, ADD, $\text{Token}_{org}$, and $\text{Token}_{wt}$ respectively represent the token with spatial positional information, matrix summation function, original tokens, and the learned spatial information weights of the 3D patch.

\begin{figure*}
\centering
\includegraphics[width=1\linewidth]{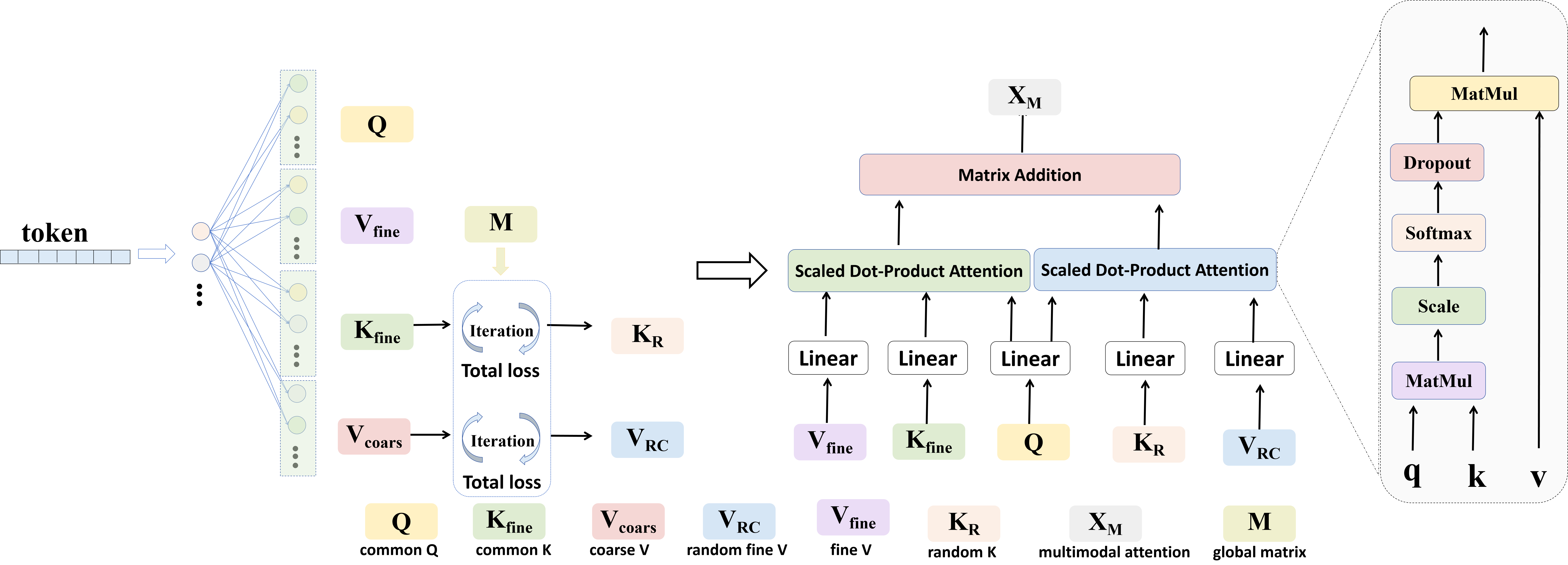}
\caption{\label{fig:qkv8}The shared query structure of the internal self-attention mechanism in SegStitch.the input feature map $x$ is fed into the fine-grained and coarse-grained attention modules of the SegStitch. The weights of the linear layers for $Q$ are shared between the two attention modules. Additionally, using the global learnable parameter matrix $M$, new parameters $\text{K}_R$ and $\text{V}_RF$ are obtained by multiplying $M$ with $\text{K}_{fine}$ and $\text{V}_{fine}$ respectively, resulting in a new set of QKV combinations. Thereby, enabling different attention tasks for each self-attention module.}
\end{figure*}

\subsection{Network Encoder}

Applying the Transformer architecture to medical imaging tasks involves a critical consideration: converting the entire image into a sequence for the self-attention mechanism. This approach can lead to issues such as excessive computational costs, significant storage requirements, and the inability to fully retrieve spatial information. To address these drawbacks, the self-attention module in the SegStitch method utilizes shared query techniques. This not only demonstrates enhanced feature learning capabilities but also effectively addresses the challenges of long-distance edge segmentation through the attention mechanism, achieving outstanding performance across multiple datasets.

\subsubsection{Shared Queries}

When constructing the 3D feature extraction module SegStitch, we adopted a novel self-attention mechanism that introduces both fine-grained and coarse-grained attention modules through shared queries. The entire image is divided into non-overlapping patches for parallel processing, reducing the computational complexity of the self-attention mechanism from quadratic to linear levels, thus achieving more efficient computation. The fine-grained attention module effectively focuses on small target features within the input sequence, while the coarse-grained attention module effectively establishes dependencies between different positions in the input sequence.
As shown in Figure \ref{fig:qkv8}, The computations for the two attention modules are as follows:

\begin{equation}
\text{Attn}_f = \text{FA}(\text{Q}_{common}, \text{K}_{fine}, \text{V}_{fine})
\end{equation}

\begin{equation}
\text{K}_{R},\text{V}_{RC} = \text{K}_{fine} \cdot \text{RP}(\text{M}) , \text{V}_{coarse} \cdot \text{RP}(\text{M})
\label{eq:3}
\end{equation}

\begin{equation}
\text{Attn}_c = \text{CA}(\text{Q}_{common}, \text{K}_{R}, \text{V}_{RC})
\end{equation}

Where $\text{Attn}_f$ and $\text{Attn}_c$ respectively represent the fine-grained and coarse-grained attention maps. In this system or module, $\text{FA}$ denotes the fine-grained attention module, while $\text{CA}$ denotes the coarse-grained attention module, $\text{RP}$ represents the random learnable parameter module, and $\text{M}$ is a two-dimensional matrix initialized to all zeros. $Q_{\text{common}}$, $K_{\text{fine}}$, $\text{K}_R$, $V_{\text{fine}}$, and $V_{\text{RC}}$ are matrices used for shared queries, fine-grained keys, coarse-grained keys, fine-grained value layers, and coarse-grained value layers respectively.

Fine-grained Attention: Given a normalized tensor $X$ of shape HWD × C, we use three linear layers to compute projections for $\text{Q}_{{common}}$, $\text{K}_{{fine}}$, and $\text{V}_{{fine}}$, obtaining $\text{Q}_{{common}} = \text{W}_QX$, $\text{K}_{{fine}} = \text{W}_KX$, and $\text{V}_{{fine}} = \text{W}_VX$, where X is the input feature map, and $\text{W}_Q$, $\text{W}_K$, and $\text{W}_V$ are the projection weights for $\text{Q}_{{common}}$, $\text{K}_{{fine}}$, and $\text{V}_{{fine}}$ respectively, with sizes H×W×D×C. 
The fine-grained attention is defined as follows:

\begin{equation}
\text{Attn}_f = \text{Softmax}\left(\frac{Q_{\text{common}}\text{K}_{{fine}}^T}{\sqrt{d}}\right) \cdot \tilde{\text{V}}_{\text{fine}}
\end{equation}

Where $\text{Q}_{{common}}$, $\text{K}_{{fine}}$, and $\text{V}_{{fine}}$ respectively represent the shared query, the projected shared key, and the projected fine-grained feature value layer, and d is the size of each vector.

Coarse-grained Attention: In this module, we inherit $\text{Q}_{common}$ from the fine-grained attention module and combine it with $\text{K}_R$ and $\text{V}_{RC}$ generated by Formula \ref{eq:3} to form a new attention mechanism. This approach enhances the ability to capture dependencies between different positions within the input sequence. Coarse-grained attention is defined as follows:

\begin{equation}
\text{Attn}_c =   \text{Softmax}\left(\frac{\text{Q}_{common}^T \text{K}_{R}}{\sqrt{d}}\right) \cdot \text{V}_{RC}
\end{equation}

Where $\text{V}_{{RC}}$, $\text{Q}_{{common}}$, and $\text{K}_{{R}}$ respectively represent the coarse-grained value layer, the shared query, and the coarse-grained key, and d is the size of each input token.

\begin{figure}
\centering
\includegraphics[width=1\linewidth]{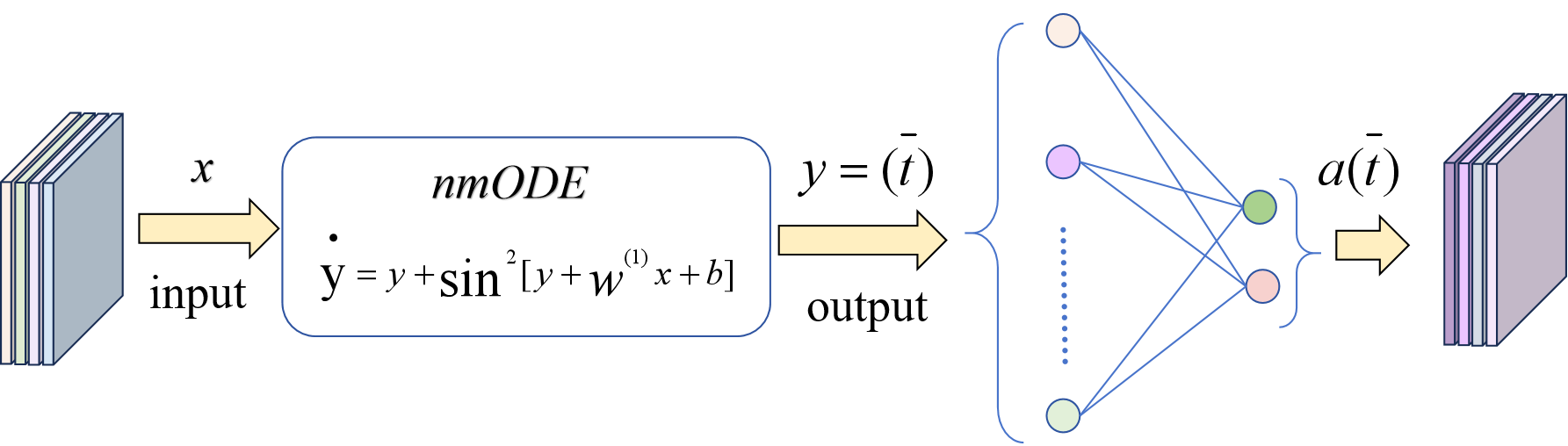}
\caption{\label{fig:nmODE}Structure diagram of nmODE. By introducing the ODE block in SegStitch, we have achieved a more stable feature representation without sacrificing recognition ability.}
\end{figure}
\subsubsection{nmODE}

Yi \cite{35} proposed a novel type of neural memory ordinary differential equation, the nmODE, which overcomes the limitations of traditional differential equations while fully leveraging the storage capacity provided by dynamical systems. Therefore, the nmODE module can simultaneously learn and memorize dependencies between characteristics during the mapping process. For individual samples, the high nonlinearity and memory capacity of the nmODE more effectively refine and integrate feature information from different channels, thereby obtaining more global feature information.

\begin{equation}
\dot{\text{y}} = \text{-y} + \sin^2(\text{y}(\text{t}) + \text{W}^{(1)}\text{x} + \text{b})
\end{equation}

Where $y \in \mathbb{R}^{n}$, $x \in \mathbb{R}^{m}$, $\text{b} \in \mathbb{R}^{n \times n}$, and $\text{W}^{(1)} = \text{W}_{ij}^{(1)} \times m \in \mathbb{R}^{n \times m}$. From the perspective of neural networks, x represents the external input of the feature map, $\text{y}(\text{t})$ represents the state of the network at time t. $\text{W}$ and $\text{b}$ respectively represent the learned weights and biases.

In this study, an nmODE module is embedded within the decoder to further optimize the model's upsampling and skip connections, fully integrating feature information from different channels. This leverages the powerful non-linear expressive capabilities of nmODE, enhancing the model's ability to represent and learn in complex scenarios. This integration is illustrated in Figure \ref{fig:nmODE}.In summary, the final output $\hat{\text{X}}$ is expressed as:

\begin{equation}
\text{X}_{out} = \text{nmODE}(\text{Attn}_f + \text{Attn}_c)
\end{equation}

where $\text{Attn}_f$ and $\text{Attn}_c$ represent the coarse-grained and fine-grained attention maps, respectively, and nmODE is the ordinary differential neural network.

\begin{table}[]
\vspace{-10pt} 
\begin{center}
\setlength{\tabcolsep}{4pt} %
\centering
\caption{Details of the medical image datasets used for evaluation.}
\begin{tabular}{cccc}
\hline
\textbf{Dataset} & \textbf{Image resolution range (mm\textsuperscript{3})} & \textbf{Format} & \textbf{Classes} \\ \hline
ACDC & ([0.7$\sim$1.95]×[0.7$\sim$1.95]×[5.0$\sim$10.0]) & 16x160x160 & 3 \\ %
Synapse & ([0.54$\sim$0.54]×[0.98$\sim$0.98]×[2.5$\sim$5.0]) & 64x128x128 & 8 \\ \hline
\end{tabular}
\label{tab:dataset_2}
\end{center}
\vspace{-10pt} 
\end{table}

\section{ Experiments and results}
\subsection{Dataset}

Our network underwent rigorous evaluation on two formidable medical image datasets: the Synapse Multi-organ Segmentation Dataset (Synapse) \cite{19} and the Automatic Cardiac Diagnosis Challenge (ACDC) \cite{20}. A comprehensive summary of the datasets' characteristics is delineated in Table \ref{tab:dataset} and Table \ref{tab:dataset_2}. These datasets are publicly accessible, offering a robust framework for the assessment of image segmentation methodologies against ground-truth (GT) masks.

\begin{table}[]
\setlength{\tabcolsep}{10pt} %
\renewcommand{\arraystretch}{1.5} %
\begin{center}
\caption{ACDC and Synapse Experimental Dataset Distribution.}
\begin{tabular}{
>{\columncolor[HTML]{FFFFFF}}l 
>{\columncolor[HTML]{FFFFFF}}l 
>{\columncolor[HTML]{FFFFFF}}l 
>{\columncolor[HTML]{FFFFFF}}l 
>{\columncolor[HTML]{FFFFFF}}l }
\hline
\cellcolor[HTML]{FFFFFF}                                   & \multicolumn{4}{c}{\cellcolor[HTML]{FFFFFF}\textbf{Image count}} \\ \cline{2-5} 
\multirow{-2}{*}{\cellcolor[HTML]{FFFFFF}\textbf{dataset}} & Training        & Validation       & Testing       & Total       \\ \hline
ACDC                                                       & 70              & 10               & 20            & 100         \\ 
Synapse                                                    & 18              & -                & 12            & 30          \\ \hline
\end{tabular}
\label{tab:dataset}
\end{center}
\end{table}

(1)The Automatic Cardiac Diagnosis Challenge (ACDC) dataset comprises 100 three-dimensional cardiac magnetic resonance imaging (MRI) cases collected from diverse patients, encompassing the cardiac organ from the base of the left ventricle to the apex. The series offers a wide range of slice thicknesses (from 5.0mm to 10.0mm) and in-plane resolutions (from 0.7x0.95mm$^{2}$ to 0.7x1.95mm$^{2}$). For the experiment at hand, the dataset has been partitioned into training, validation, and testing sets in a 7:1:2 ratio. To ensure fair comparison with previous research methodologies, images have been randomly cropped to a model input size of 160x160x16, in accordance with the standard procedure.

(2)In our research, we utilized the Synapse Multi-Organ Segmentation Dataset, a publicly available resource that originated from the 2015 MICCAI Multi-Atlas Abdominal Labeling Challenge. The dataset encompasses 30 abdominal computed tomography (CT) scans, demonstrating a range of slice thicknesses from 2.5mm to 5.0mm, as well as varying in-plane resolutions from 0.54x0.54mm² to 0.98x0.98mm². With meticulous attention to detail, we meticulously divided the dataset into a training set comprising 18 cases and a testing set comprising 12 cases. To ensure a fair and comparable evaluation against previous research methodologies, all images were uniformly cropped to a size of 128x128x64 voxels during the inference phase, tailored to the model's input specifications.

\subsection{Evaluation criteria}

HD95 is a metric used to assess the difference between two point sets. It focuses on the closest 95\% of points between the predicted and the ground truth segmentation, excluding the most extreme 5\% to reduce the impact of outliers. This method provides a reliable measure of segmentation accuracy and reflects the model's performance more realistically. The Dice Similarity Coefficient (mDSC) is used to quantify the similarity between the predicted and ground truth segmentations in medical images. By using both HD95 and mDSC metrics simultaneously, we can present a more comprehensive view of the model's performance and accuracy.

\subsection{Training details}

Pretraining is crucial for Transformer-based models, such as Vision Transformers (ViT) \cite{8}, where the model's performance is widely acknowledged to be heavily reliant on pretraining, a notion corroborated by empirical evaluations. Established medical image segmentation techniques also commonly leverage pretrained weights to initialize their models \cite{50},\cite{46},\cite{22},\cite{53}. However, pretraining Transformer-based models presents two challenges. Firstly, it is computationally intensive, often demanding substantial resources. Secondly, unlike the natural scene images in ImageNet \cite{54}, the medical imaging field lacks large-scale datasets for pretraining. Consequently, our SegStitch model opted for training from scratch, meaning we randomly initialized the model weights. Despite this, our experiments on two major public datasets, ACDC and Synapse, yielded promising performance comparable to state-of-the-art pretrained methods.

\begin{table}[]
\begin{center}

\caption{Ablation Studies on self-attention and ODE block in SegStitch Segmentation.The best scores are indicated in \textbf{bold}.}
\begin{tabular}{lllll}
\hline
\multicolumn{1}{c}{Methods}      & &\multicolumn{1}{c}{self-attention} & \multicolumn{1}{c}{ODE block} & \multicolumn{1}{c}{mDSC(\%)} \\ \hline
 &1& $\surd$ & $-$ & 88.3 \\
 \multirow{2}{*}{\textbf{SegStitch (Ours)}}&2& $-$ & $\surd$ & 87.8 \\
 &3& $-$ & $-$ & 87.6 \\
 &4& $\surd$ & $\surd$ & \textbf{89.8} \\ \hline
\end{tabular}
\label{tab:Ablation}
\end{center}
\end{table}

We implemented our models in PyTorch 1.8.0 and trained them on an NVIDIA GeForce RTX 3090 GPU with 24 GB memory. The batch size was 2, and training was capped at 1000 epochs. We used the SGD optimizer with a momentum of 0.99 \cite{55}.

\begin{table}[]
\begin{center}
\caption{Segmentation performance of different methods on the ACDC dataset.The best scores are indicated in \textbf{bold}.}
\begin{tabular}{l|lll|ll}
\hline
\multicolumn{1}{c}{Method}         & \multicolumn{1}{c}{RV$^{\uparrow}$}    & \multicolumn{1}{c}{Myo}   & \multicolumn{1}{c}{LV$^{\uparrow}$}    & \multicolumn{1}{c}{AVG$^{\uparrow}$}

\\ \hline
R50 U-Net\cite{38}      & 87.10 & 80.63 & 94.92 & 87.55    \\
R50 Att-UNet\cite{39}   & 87.58 & 79.20 & 93.47 & 86.75   \\
VIT\cite{8}            & 81.46 & 70.71 & 92.18 & 81.45   \\
R50 VIT\cite{8}        & 86.07 & 81.88 & 94.75 & 87.57   \\
TransUNet\cite{43}      & 88.86 & 84.54 & 95.73 & 89.71   \\
Swin-UNet\cite{14}      & 88.55 & 85.62 & 95.83 & 90.00   \\
LeVit-Unet-384\cite{46} & 89.55 & 87.64 & 93.76 & 90.32   \\
nnFormer\cite{22}       & 90.22 & 89.53 & 95.59 & 91.78   \\
MISSFormer\cite{48}     & 86.36 & 85.75 & 91.59 & 87.90   \\
D-Former\cite{23}       & 91.33 & 89.60 & 95.93 & 92.29   \\
UNETR\cite{21}          & 85.29 & 86.52 & 94.02 & 86.61   \\
UNETR++\cite{29}        & \underline{91.89} & \textbf{90.61} & \underline{96.00} & \underline{92.83}    \\
\rowcolor[HTML]{FFF2CC} 
\textbf{SegStitch (Ours)}           & \textbf{93.05} & \underline{90.46} & \textbf{96.46} & \textbf{93.32}   \\ \hline
\end{tabular}
\label{tab:ACDC}
\end{center}
\vspace{-10pt} 
\end{table}

\begin{table*}[]
\begin{center}
\caption{Segmentation performance of different methods on the Synapse dataset.The best scores are indicated in \textbf{bold}.}
\small %
\begin{tabular}{l|llllllll|ll}
\hline
\multicolumn{1}{c}{Method}                                             & \multicolumn{1}{c}{Aor$^{\uparrow}$} & \multicolumn{1}{c}{Gal} & \multicolumn{1}{c}{LKid$^{\uparrow}$} & \multicolumn{1}{c}{RKid$^{\uparrow}$} & \multicolumn{1}{c}{Liv} & \multicolumn{1}{c}{Pan} & \multicolumn{1}{c}{Spleen} & \multicolumn{1}{c}{Sto$^{\uparrow}$} & 
\multicolumn{1}{c}{AVG$^{\uparrow}$} &
\multicolumn{1}{c}{HD95$^{\downarrow}$} 
\\ \hline
V-Net\cite{36}                                      & 75.34 & 51.87       & 77.10      & 80.75      & 87.84 & 40.04    & 80.56  & 56.98   & 68.81    &-                                          \\
DARR\cite{2}                                       & 74.74 & 53.77       & 72.31      & 73.24      & 94.08 & 54.18    & 89.90  & 45.96   & 69.77    &-                                          \\
R50 U-Net\cite{2}                                  & 87.74 & 63.66       & 80.60      & 78.19      & 93.74 & 56.90    & 85.87  & 74.16   & 74.68    &-                                          \\
R50 Att-UNet\cite{39}                               & 55.92 & 63.91       & 79.20      & 72.71      & 93.56 & 49.37    & 87.19  & 74.95   & 75.57    &-                                          \\
U-Net\cite{2}                                      & 89.07 & 69.72       & 77.77      & 68.60      & 93.43 & 53.98    & 86.67  & 75.58   & 76.85    &39.7                                          \\
Att-UNet\cite{41}                                   & 89.55 & 68.88       & 77.98      & 71.11      & 93.57 & 58.04    & 87.30  & 75.75   & 77.77    &36.02                                          \\
VIT\cite{8}                                        & 70.19 & 45.10       & 74.70      & 67.40      & 91.32 & 42.00    & 81.75  & 70.44   & 67.86    &36.11                                          \\
R50 VIT\cite{8}                                    & 73.73 & 55.13       & 75.80      & 72.20      & 91.51 & 45.99    & 81.99  & 73.95   & 71.29    &32.87                                          \\
TransUNet\cite{43}                                  & 87.23 & 63.13       & 81.87      & 77.02      & 94.08 & 55.86    & 85.08  & 75.62   & 77.48    &31.84                                          \\
Swin-UNet\cite{14}                                  & 85.47 & 66.53       & 83.28      & 79.61      & 94.29 & 56.58    & 90.66  & 76.60   & 79.13    &22.96                                          \\
nnFormer\cite{22}                                   & 92.04 & 71.09       & 87.64      & 87.34      & 96.53 & \textbf{82.49}    & 92.91  & \underline{89.17}   & 87.40    &10.63                                          \\
MISSFormer\cite{48}                                 & 86.99 & 68.65       & 85.21      & 82.00      & 94.41 & 65.67    & 91.92  & 80.81   & 81.96   &18.20                                           \\
D-Former\cite{23}                                   & 92.12 & \textbf{80.09}       & \underline{92.60}      & \underline{91.91}      & \textbf{96.99} & 76.67    & \underline{93.78}  & 86.44   & \underline{88.83}     &-                                         \\
UNETR++\cite{29}                                    & \underline{92.52} & 71.25       & 87.54      & 87.18      & 96.42 & \underline{81.10}    & \textbf{95.77}  & 86.01   & 87.22     &\textbf{7.53}                                         \\ \hline
\rowcolor[HTML]{FFF2CC} 
\multicolumn{1}{|l|}{\cellcolor[HTML]{FFF2CC}\textbf{SegStitch (Ours)}} & \textbf{93.61} & \underline{77.11}       & \textbf{94.17}      & \textbf{92.86}      & \underline{96.92} & 80.39    & 92.61  & \textbf{90.79}   & \multicolumn{1}{l}{\cellcolor[HTML]{FFF2CC}\textbf{89.80}} & {8.48}  \\ \hline
\end{tabular}
\label{tab:Synapse}
\end{center}
\vspace{-10pt} 
\end{table*}

\subsection{ Results and comparisons with recent methods}

In this section, the performance of the proposed SegStitch is compared with state-of-the-art methods on the ACDC and Synapse datasets.

\begin{table}[]
\begin{center}
\caption{Computational complexity analysis of the proposed SegStitch. All comparisons were conducted on the Synapse dataset.The best scores are indicated in \textbf{bold}.}
\begin{tabular}{l|ll|l}
\hline
\textbf{Method}     & \textbf{Params(M)} & \textbf{GFLOPs} & \textbf{mDSC(\%)} \\ \hline
{TransUNet}  & 96.07              & 88.91           & 77.48            \\
{UNETR}      & 92.49              & 75.46           & 78.35            \\
{Swin-UNETR} & 62.83              & 384.2           & 83.48            \\
{nnFormer}   & 150.5              & 213.4           & 86.57            \\ \hline
{\textbf{SegStitch (Ours)}}       & \textbf{60.81}              & \textbf{67.36}           & \textbf{89.80}            \\ \hline
\end{tabular}
\label{tab:complexity}
\end{center}
\end{table}

 \subsubsection{Comparisons with state-of-the-art methods on ACDC datase}

We evaluated the performance of our proposed SegStitch model on the ACDC dataset and compared it with various state-of-the-art models. The segmentation results on the ACDC dataset are presented in Table \ref{tab:ACDC}, from which relevant conclusions can be drawn. SegStitch achieved the best average Dice Similarity Coefficient (mDSC) of 93.32\% without pre-training. Compared to the R50 U-Net and R50 Att-UNet methods, our approach resulted in average mDSC improvements of 5.77\% and 6.57\%, respectively. When compared to methods based on concurrent Transformers, our approach still obtained performance gains of 0.45\% and 1.54\% in average mDSC over UNetr++ and nnFormer, respectively. Specifically, SegStitch demonstrated superior segmentation accuracy across all crucial cardiac components, including the Right Ventricle (RV), Myocardium (Myo), and Left Ventricle (LV).

\subsubsection{Comparisons with state-of-the-art methods on Synapse datase}

We conducted experiments comparing the Synapse multi-organ segmentation dataset with state-of-the-art methods, as shown in Table \ref{tab:Synapse}. Firstly, it can be observed that traditional CNN methods still exhibit good performance. However, the addition of Transformers or the use of pure Transformer architectures has been shown to have significant effectiveness, surpassing CNNs to some extent. Among these frameworks, R50 U-Net, R50 Att-Unet, and R50 ViT achieved mDSC scores of over 70\% compared to V-Net, DARR, and ViT, indicating the overall superiority of Transformer-based models over traditional CNN models.

In our comparison with the UNETR++ model, which integrates a Transformer into a multi-attention mechanism, our experimental results demonstrate that our approach has achieved superior segmentation performance. Specifically, it has achieved an average accuracy of 89.80\% on the Synapse dataset, along with an HD95 metric of 8.48. It is evident that our algorithm achieves a 2.58\% improvement in average accuracy. Our model outperforms a range of CNN-based approaches (e.g., R50 U-Net, R50 Att-Unet, U-Net, Att-Unet) and Transformer-based methods (e.g., R50 ViT, TransUnet, Swin-Unet), including the Transformer-integrated UNETR++. It not only delivers superior overall segmentation results but also achieves the highest performance in segmenting specific organs such as the Aorta, Kidney(L), Kidney(R), and Stomach, surpassing the next best by margins of 1.09\%, 1.57\%, 0.95\%, and 1.62\%, respectively.

\begin{figure}
\vspace{-10pt} 
\centering
\includegraphics[width=1\linewidth]{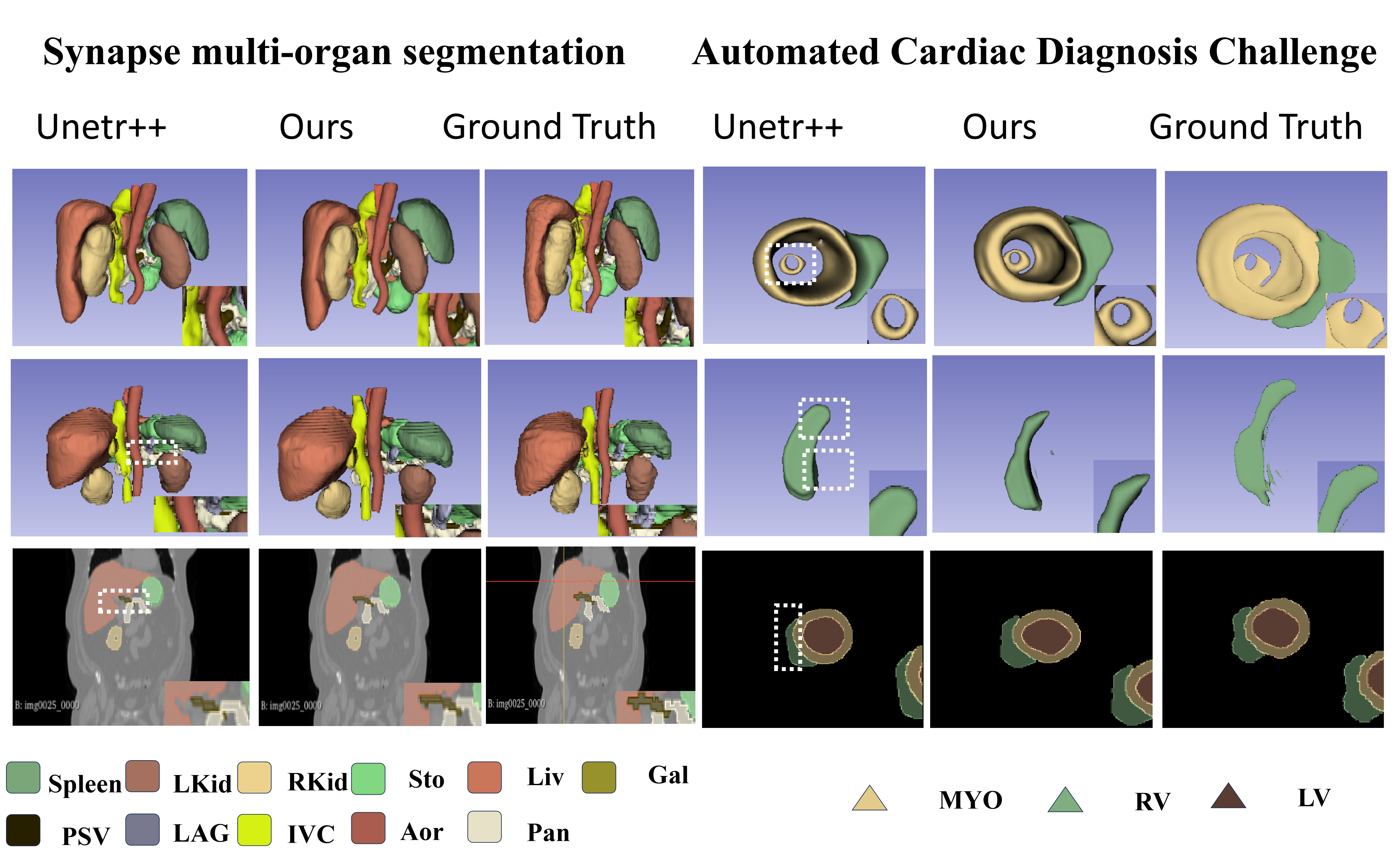}
\caption{\label{fig:show7}Qualitative validation and comparison of segmentation results from different methods on the Synapse multi-organ dataset and the ACDC cardiac segmentation dataset. From left to right: UNETR++, SegStitch (Ours), Ground Truth. Comprehensive analysis shows that we achieve finer and more accurate segmentation results.}
\vspace{-10pt} 
\end{figure}

The experimental results further validate the feasibility of our proposed Transformer-based multi-attention approach. Figure \ref{fig:show7} also demonstrates some segmentation results. From these images, it can be observed that our framework can improve segmentation accuracy to a certain extent. The SegStitch model can learn high-level semantic features and low-level texture features, achieving precise localization and segmentation.

\subsection{SegStitch computational complexity comparisons}

The SegStitch method demonstrates significant advantages in computational complexity, primarily due to the substantial reduction in the number of learnable parameters. It is impressive that, despite having only about 6.08 million parameters, SegStitch has outperformed many other algorithms. As shown in Table \ref{tab:complexity}, SegStitch has significantly improved efficiency by reducing the number of learnable parameters and the number of floating-point operations (FLOPs), yet it still maintains a first-class performance compared to other methods. SegStitch achieves a perfect balance between the number of parameters and excellent segmentation results, proving its efficiency in 3D medical image segmentation tasks. Furthermore, with only 67.36 giga floating-point operations, the compactness of SegStitch greatly simplifies the deployment and application of this method in actual clinical settings. The reduction in model size not only enhances the efficiency of medical imaging analysis but also strengthens its effectiveness in practical applications, thereby facilitating seamless integration in clinical practice.

\subsection{Ablation study of Synapse dataset}

We conducted an in-depth ablation study on the core components of SegStitch using the Synapse dataset, aiming to explore the effectiveness of the proposed self-attention mechanism and to assess the specific impact of image input size on the model's segmentation accuracy. In our experimental design, we deliberately removed the self-attention mechanism or ODE block to verify the contribution of these key components to overall performance. The results are summarized in Table \ref{tab:Ablation}. The findings indicate that both the self-attention mechanism and ODE block are crucial for maintaining the model's high performance, with the absence of either significantly degrading the model's performance. Overall, SegStitch achieves superior segmentation performance through its expanded attention mechanism and self-learning structure. The experiments further emphasize the importance of information exchange between samples, with the absence of self-attention mechanism leading to increased instability in the results. The ODE block, on the other hand, exhibits lower variability in results, which we speculate may be due to the global perception capabilities of the Transformer architecture, significantly enhancing the model's robustness.

\subsection{Discussion}
\subsubsection{clinical impact}

Automated localization and segmentation of the heart and abdominal organs in MRI are crucial for various clinical applications, as they directly impact subsequent quantitative analysis and diagnosis. Although fully automated systems have not yet fully integrated into routine clinical practice, developing robust frameworks can significantly enhance diagnostic workflows. For instance, cardiac MRI segmentation is essential for measuring blood pool and heart volumes, which helps derive functional indices such as ventricular volume and ejection fraction. These quantification metrics are used for patient diagnosis. Therefore, this paper proposes an end-to-end network for automatic segmentation of the heart and abdominal organs. The network effectively delineates cardiac structures and, due to its lightweight design and long-range segmentation capabilities, offers significant advantages in clinical settings (see Figure \ref{fig:show7}).

\subsubsection{Performance Analysis of the Self-Attention Mechanism}

\begin{figure}
\centering
\includegraphics[width=1\linewidth]{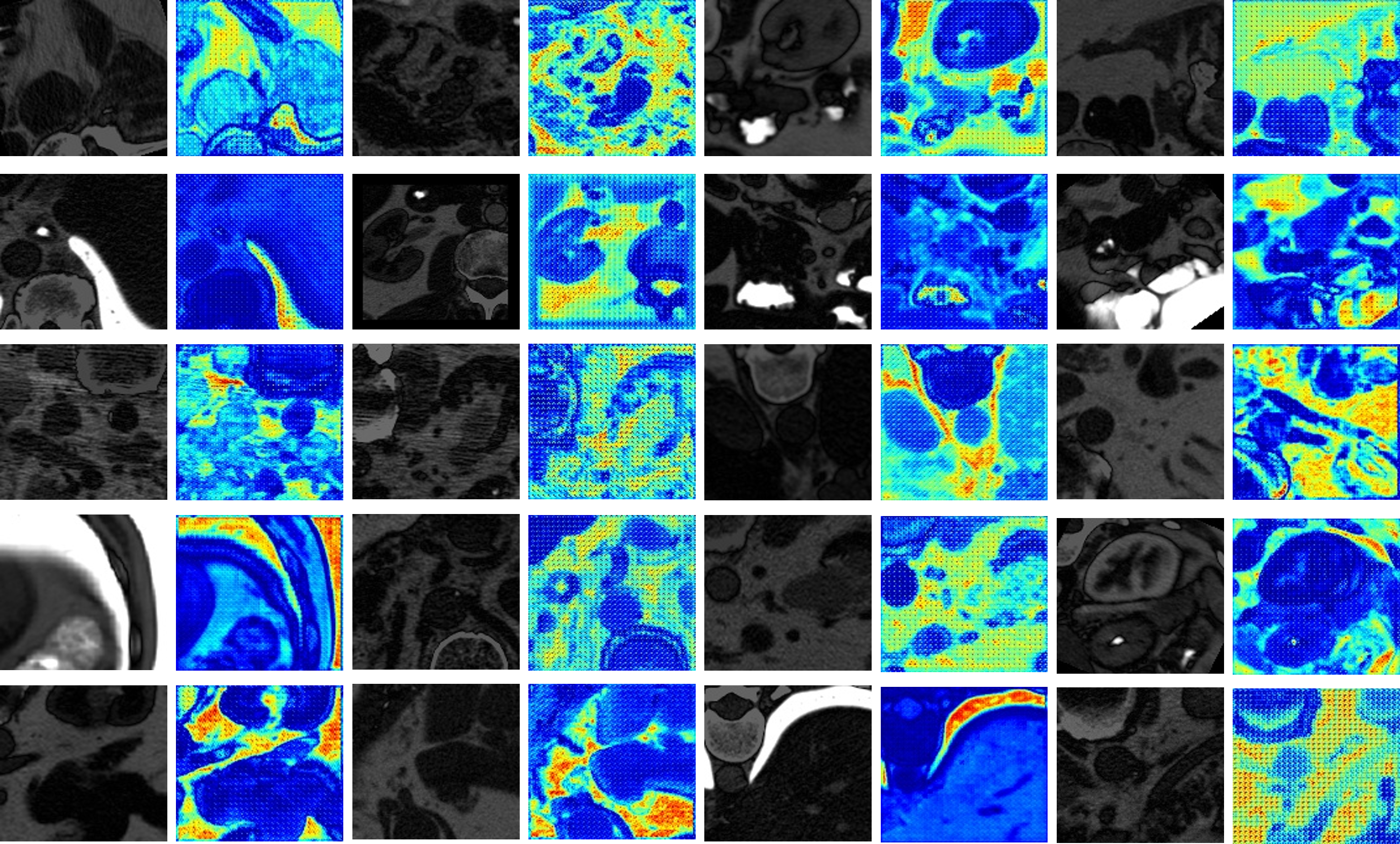}
\caption{\label{fig:hot}Visualization of weights for SegStitch self-attention in feature maps. Darker colors indicate higher weight values. This figure demonstrates the regions of focus of the self-attention on the feature maps. From the figure, it can be observed that the self-attention module pays more attention to the elongated parts in the feature maps. This is because the ViT block with spatial positional information tends to learn global features. }
\end{figure}

As shown in Figure \ref{fig:hot}, we extract the weight values of the feature maps in the self-attention module and visualize them in the form of a heatmap. experiments indicate that the self-attention module focuses more on presenting distant features. Leveraging this advantage, it significantly concentrates on accurately segmenting regions with elongated shapes. In summary, this method effectively learns global semantic information and local spatial detail features, with particularly outstanding segmentation of elongated structures.

To vividly demonstrate the performance of our SegStitch model, we conducted qualitative comparisons with UNETR++ (as shown in Figure \ref{fig:show7}). It is evident that on the Synapse dataset, our model's organ segmentation results are generally closer to reality. For specific organs, our model can accurately identify and outline the contours of the kidneys, capturing some detailed contours, which aligns with the conclusions of our experimental results. On the ACDC dataset, our model excels at delineating the contours of the right ventricle. Since the right ventricle typically exhibits a long and narrow shape, this indicates that our model possesses excellent long-range dependency capture and detail processing capabilities.

\begin{table}[]
\begin{center}
\caption{\label{tab:5_train_v_table}Comparison Analysis of the Impact of ODE block on Loss and Dice Similarity Coefficient in Different Ordinary Differential Equations and Alternative Modules on the ADCD Dataset.The best scores are indicated in \textbf{bold}.}
\begin{tabular}{@{}ccc|ccc@{}}
\hline
Basemodel             &                        & Function & train loss & validation loss & mDSC(\%) \\ \hline
\multirow{5}{*}{SegStitch} & \multirow{3}{*}{ODE} &$\sin^2$     & -0.9249      & \textbf{-0.8895}           & \textbf{93.32}   \\
                      &                        & sin       & \textbf{-0.9289}      & -0.8830           & 93.15   \\
                      &                        & tan       & -0.9287      & -0.8832           & 93.07   \\
                      & AE                     & -         & -0.9273      & -0.8753           & 92.38   \\
                      & sigmoid                & -         & -0.9275      & -0.8711           & 92.15   \\
                      & -                     & -         & -0.9191      & -0.8741           & 92.21   \\ \hline
\end{tabular}
\end{center}
\vspace{-10pt} 
\end{table}

\subsubsection{Robustness Analysis of ODE block}
Developing a method endowed with robust capabilities is of paramount importance, particularly in the dynamic field of medical image analysis. The robustness of a model is indicative of its superior ability to effectively learn from complex and novel datasets. In this section, our objective is to demonstrate the exceptional resilience of SegStitch by conducting a comprehensive evaluation of its segmentation performance on the ACDC dataset.

For the segmentation task on the ACDC dataset, we conducted a comparative analysis of the performance between the model's blocks: one version incorporating the ODE block and another without it. The comparative results are presented in Table \ref{tab:5_train_v_table}. Notably, upon the incorporation of the ODE block, the training and validation loss curves during the training process exhibited a smoother descent. This robust performance effectively addresses the complexities inherent in medical image data, underscoring the efficacy of our approach.

In addition to the aforementioned, we have conducted an investigation into the impact of various modules on the performance of SegStitch. The findings are delineated in Table \ref{tab:5_train_v_table}. It should be noted that '$\sin^2$', 'sin', and 'tanh' represent the ablation studies conducted on different ordinary differential equations within the ODE block framework. 'Sigmoid' refers to the scenario where the ODE block is supplanted by a sigmoid activation function, while 'AE' signifies the substitution of the ODE block with a more parameter-rich autoencoder. Our results indicate that the ODE block achieves optimal robustness when employing '$\sin^2$' as the ODE. Compared to the 'sigmoid' activation function, the ODE-based neural network demonstrates superior performance in data fitting. Furthermore, we conducted comparative experiments using 'AE'. During the training phase, the 'AE' experiment's training and validation losses exhibited more pronounced fluctuations, indicating that the model's performance is not solely a result of parameter stacking.

\vspace{-0.1cm}

\section{Conclusion}

In this paper, we introduce SegStitch, an innovative framework for medical image segmentation. Initially, it fuses 3D image features with spatial positional cues, thereby maintaining the integrity of spatial information within the images. Subsequently, by incorporating both coarse and fine-grained attention mechanisms into the Transformer architecture, we have markedly amplified the model's capacity to manage long-range dependencies and discern intricate details. Furthermore, the integration of an ODE block into SegStitch marks a pioneering step, significantly bolstering the model's robustness. To encapsulate, this paper unveils SegStitch, a novel framework for 3D medical image segmentation. Empirical validation demonstrates that our model not only streamlines parameter count and computational load but also surpasses existing benchmarks in segmentation performance on the public datasets Synapse and ACDC. This study heralds a significant leap in medical image analysis, with the potential to profoundly impact future medical image processing endeavors.

\appendices

\section{Ablation Study of ODE block}

\begin{figure}[H]
\vspace{-10pt} 
\centering
\includegraphics[width=0.9\linewidth]{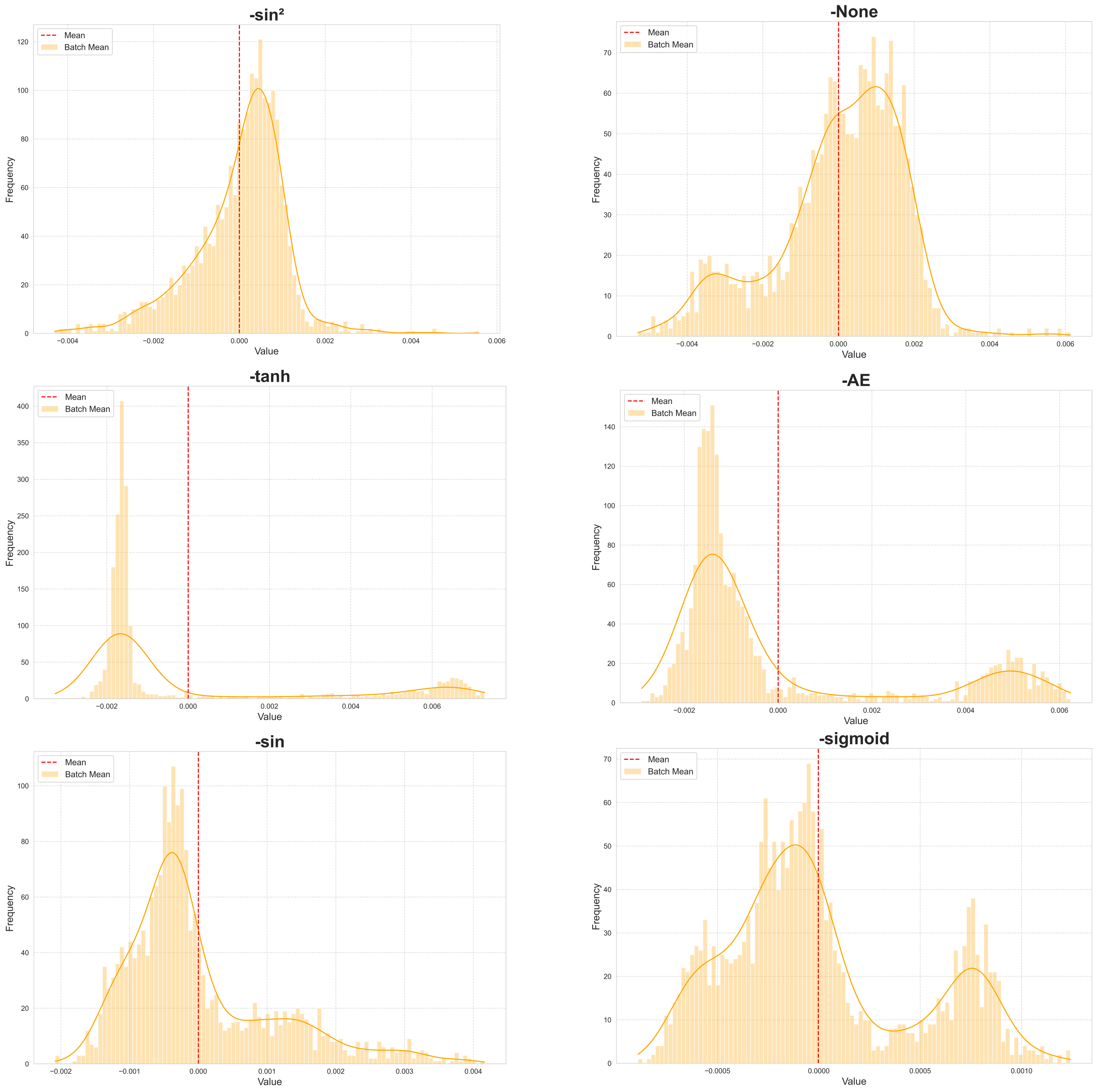}
\caption{\label{fig:soft_point}On the ADCD dataset, we perform channel mean statistics on features processed by the encoder to compare the performance of different ordinary differential equations and alternative modules.}

\end{figure}

To further investigate the rationale behind these observations, we conducted zero-mean feature experiments following the ODE block. Using 40 samples, we calculated the mean of the encoded features for each channel by the model. The results suggest that the model's feature distribution becomes more uniform with the ODE in use. Moreover, when using '$\sin^2$' as the ODE, the channel means are more concentrated and align more closely with the normal distribution, enhancing the model's convergence and stability. The experimental outcomes are depicted in Figure \ref{fig:soft_point}.

\begin{figure}[H]
\vspace{-10pt} 
\centering
\includegraphics[width=1\linewidth]{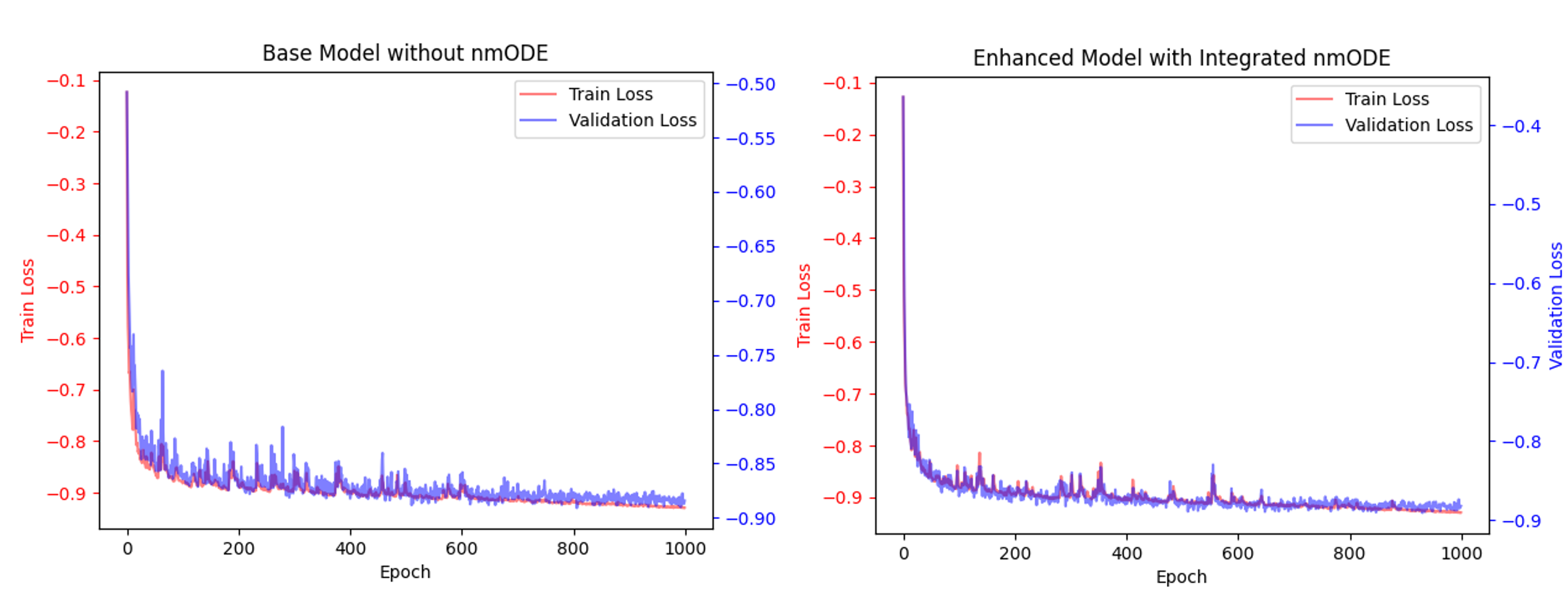}
\caption{\label{fig:N_Y_nmODE}Conducting an ablation study on the nmODE model using the ACDC dataset, with a comparative analysis of the model's training loss and validation loss throughout the training process.}
\end{figure}

We performed a comparative analysis on the ACDC dataset between models that include an ODE block and those that do not. The results indicate that, with the introduction of the ODE block, the training and validation loss curves exhibit a smoother descending trend during the training process. The comparative results are shown in Figure \ref{fig:N_Y_nmODE}. This enhanced robustness effectively addresses the inherent complexity of medical image data, further validating the effectiveness of the ODE block in our model.

\begin{figure}[H]
\vspace{-10pt} 
\centering
\includegraphics[width=1\linewidth]{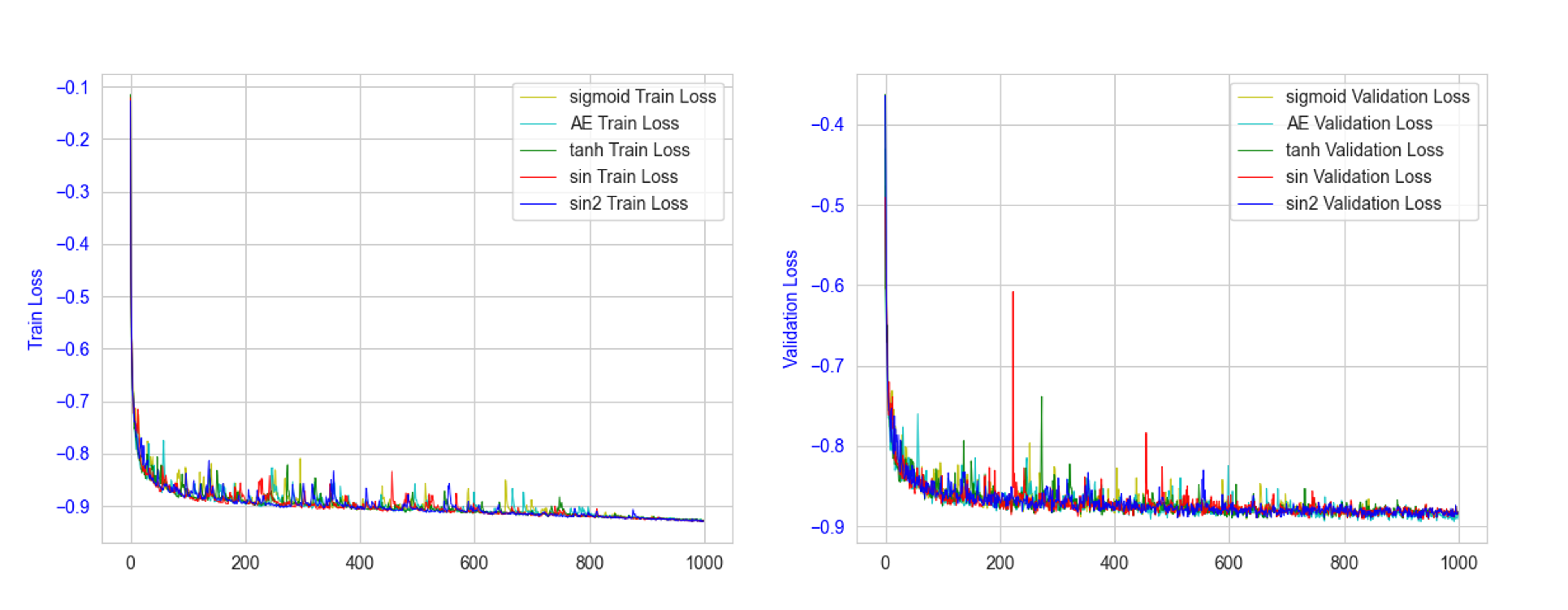}
\caption{\label{fig:5_train_v}Comprehensive Analysis of Loss Performance for the nmODE Module under Various Ordinary Differential Equations and Alternative Modules.}
\end{figure}

In addition to the previously discussed aspects, we also investigated the impact of different modules on the performance of SegStitch. The results of this study are detailed in Figure \ref{fig:5_train_v}. The terms '$\sin^2$', 'sin', and 'tanh' refer to our ablation studies on different ordinary differential equations (ODEs) used in the ODE block. 'Sigmoid' indicates that in this case, the ODE block was replaced with a sigmoid activation function, while 'AE' denotes that the ODE block was replaced with a more parameter-rich autoencoder. Our results show that using '$\sin^2$' as the ODE provided the best robustness for the ODE block module. Compared to the 'sigmoid' activation function, the ordinary differential equation neural network demonstrated superior performance in data fitting. Furthermore, we conducted comparative experiments with 'AE'. During the training phase, the training and validation losses for the 'AE' experiments exhibited more pronounced fluctuations, confirming that the model's performance issues were not merely due to parameter stacking.


\begin{thebibliography}{10}
\expandafter\ifx\csname url\endcsname\relax
  \def\url#1{\texttt{#1}}\fi
\expandafter\ifx\csname urlprefix\endcsname\relax\def\urlprefix{URL }\fi
\expandafter\ifx\csname href\endcsname\relax
  \def\href#1#2{#2} \def\path#1{#1}\fi

\bibitem{56}
Y.~Wang, Y.~Zhou, W.~Shen, S.~Park, E.~K. Fishman, A.~L. Yuille, Abdominal multi-organ segmentation with organ-attention networks and statistical fusion, Medical image analysis 55 (2019) 88--102.

\bibitem{1}
J.~Long, E.~Shelhamer, T.~Darrell, Fully convolutional networks for semantic segmentation, in: Proceedings of the IEEE conference on computer vision and pattern recognition, 2015, pp. 3431--3440.

\bibitem{2}
O.~Ronneberger, P.~Fischer, T.~Brox, U-net: Convolutional networks for biomedical image segmentation, in: Medical Image Computing and Computer-Assisted Intervention--MICCAI 2015: 18th International Conference, Munich, Germany, October 5-9, 2015, Proceedings, Part III 18, Springer, 2015, pp. 234--241.

\bibitem{3}
Z.~Zhou, M.~M. Rahman~Siddiquee, N.~Tajbakhsh, J.~Liang, Unet++: A nested u-net architecture for medical image segmentation, in: Deep Learning in Medical Image Analysis and Multimodal Learning for Clinical Decision Support: 4th International Workshop, DLMIA 2018, and 8th International Workshop, ML-CDS 2018, Held in Conjunction with MICCAI 2018, Granada, Spain, September 20, 2018, Proceedings 4, Springer, 2018, pp. 3--11.

\bibitem{4}
H.~Huang, L.~Lin, R.~Tong, H.~Hu, Q.~Zhang, Y.~Iwamoto, X.~Han, Y.-W. Chen, J.~Wu, Unet 3+: A full-scale connected unet for medical image segmentation, in: ICASSP 2020-2020 IEEE international conference on acoustics, speech and signal processing (ICASSP), IEEE, 2020, pp. 1055--1059.

\bibitem{5}
X.~Xiao, S.~Lian, Z.~Luo, S.~Li, Weighted res-unet for high-quality retina vessel segmentation, in: 2018 9th international conference on information technology in medicine and education (ITME), IEEE, 2018, pp. 327--331.

\bibitem{6}
{\"O}.~{\c{C}}i{\c{c}}ek, A.~Abdulkadir, S.~S. Lienkamp, T.~Brox, O.~Ronneberger, 3d u-net: learning dense volumetric segmentation from sparse annotation, in: Medical Image Computing and Computer-Assisted Intervention--MICCAI 2016: 19th International Conference, Athens, Greece, October 17-21, 2016, Proceedings, Part II 19, Springer, 2016, pp. 424--432.

\bibitem{7}
H.~K. Putra, B.~Suprihatin, F.~Ramadhini, et~al., ybrid clahe-gamma adjustment and densely connected u-net for retinal blood vessel segmentation using augmentation data., Engineering Letters 30~(2) (2022).

\bibitem{58}
C.~Yu, J.~Wang, C.~Gao, G.~Yu, C.~Shen, N.~Sang, Context prior for scene segmentation, in: Proceedings of the IEEE/CVF conference on computer vision and pattern recognition, 2020, pp. 12416--12425.

\bibitem{8}
A.~Dosovitskiy, L.~Beyer, A.~Kolesnikov, D.~Weissenborn, X.~Zhai, T.~Unterthiner, M.~Dehghani, M.~Minderer, G.~Heigold, S.~Gelly, et~al., An image is worth 16x16 words: Transformers for image recognition at scale, arXiv preprint arXiv:2010.11929 (2020).

\bibitem{ji2024sine}
Y.~Ji, H.~Saratchandran, C.~Gordon, Z.~Zhang, S.~Lucey, Sine activated low-rank matrices for parameter efficient learning, arXiv preprint arXiv:2403.19243 (2024).

\bibitem{wu2024xlip}
B.~Wu, Y.~Xie, Z.~Zhang, M.~H. Phan, Q.~Chen, L.~Chen, Q.~Wu, Xlip: Cross-modal attention masked modelling for medical language-image pre-training, arXiv preprint arXiv:2407.19546 (2024).

\bibitem{zhang2024jointvit}
Z.~Zhang, X.~Qi, M.~Chen, G.~Li, R.~Pham, A.~Zuhair, E.~Berry, Z.~Liao, O.~Siggs, R.~Mclaughlin, et~al., Jointvit: Modeling oxygen saturation levels with joint supervision on long-tailed octa, arXiv preprint arXiv:2404.11525 (2024).

\bibitem{9}
A.~Vaswani, N.~Shazeer, N.~Parmar, J.~Uszkoreit, L.~Jones, A.~N. Gomez, {\L}.~Kaiser, I.~Polosukhin, Attention is all you need, Advances in neural information processing systems 30 (2017).

\bibitem{10}
R.~Shao, Z.~Shi, J.~Yi, P.-Y. Chen, C.-J. Hsieh, On the adversarial robustness of vision transformers, arXiv preprint arXiv:2103.15670 (2021).

\bibitem{12}
A.~Radford, K.~Narasimhan, T.~Salimans, I.~Sutskever, et~al., Improving language understanding by generative pre-training (2018).

\bibitem{18}
H.~Yan, J.~Du, V.~Y. Tan, J.~Feng, On robustness of neural ordinary differential equations, arXiv preprint arXiv:1910.05513 (2019).

\bibitem{35}
Z.~Yi, nmode: neural memory ordinary differential equation, Artificial Intelligence Review (2023) 1--36.

\bibitem{19}
O.~Bernard, A.~Lalande, C.~Zotti, F.~Cervenansky, X.~Yang, P.-A. Heng, I.~Cetin, K.~Lekadir, O.~Camara, M.~A.~G. Ballester, et~al., Deep learning techniques for automatic mri cardiac multi-structures segmentation and diagnosis: is the problem solved?, IEEE transactions on medical imaging 37~(11) (2018) 2514--2525.

\bibitem{20}
B.~Landman, Z.~Xu, J.~Igelsias, M.~Styner, T.~Langerak, A.~Klein, Miccai multi-atlas labeling beyond the cranial vault--workshop and challenge, in: Proc. MICCAI Multi-Atlas Labeling Beyond Cranial Vault—Workshop Challenge, Vol.~5, 2015, p.~12.

\bibitem{21}
A.~Hatamizadeh, Y.~Tang, V.~Nath, D.~Yang, A.~Myronenko, B.~Landman, H.~R. Roth, D.~Xu, Unetr: Transformers for 3d medical image segmentation, in: Proceedings of the IEEE/CVF winter conference on applications of computer vision, 2022, pp. 574--584.

\bibitem{isensee2021nnu}
F.~Isensee, P.~F. Jaeger, S.~A. Kohl, J.~Petersen, K.~H. Maier-Hein, nnu-net: a self-configuring method for deep learning-based biomedical image segmentation, Nature methods 18~(2) (2021) 203--211.

\bibitem{wu2023bhsd}
B.~Wu, Y.~Xie, Z.~Zhang, J.~Ge, K.~Yaxley, S.~Bahadir, Q.~Wu, Y.~Liu, M.-S. To, Bhsd: A 3d multi-class brain hemorrhage segmentation dataset, in: International Workshop on Machine Learning in Medical Imaging, Springer, 2023, pp. 147--156.

\bibitem{zhangthin}
Z.~Zhang, B.~Zhang, A.~Hiwase, C.~Barras, F.~Chen, B.~Wu, A.~J. Wells, D.~Y. Ellis, B.~Reddi, A.~W. Burgan, et~al., Thin-thick adapter: Segmenting thin scans using thick annotations.

\bibitem{zhang2023segreg}
Z.~Zhang, X.~Qi, B.~Zhang, B.~Wu, H.~Le, B.~Jeong, M.-S. To, R.~Hartley, Segreg: Segmenting oars by registering mr images and ct annotations, arXiv preprint arXiv:2311.06956 (2023).

\bibitem{59}
N.~Ibtehaz, M.~S. Rahman, Multiresunet: Rethinking the u-net architecture for multimodal biomedical image segmentation, Neural networks 121 (2020) 74--87.

\bibitem{60}
J.~Chen, Y.~Lu, Q.~Yu, X.~Luo, E.~Adeli, Y.~Wang, L.~Lu, A.~L. Yuille, Y.~Zhou, Transunet: Transformers make strong encoders for medical image segmentation, arXiv preprint arXiv:2102.04306 (2021).

\bibitem{61}
R.~Azad, M.~T. Al-Antary, M.~Heidari, D.~Merhof, Transnorm: Transformer provides a strong spatial normalization mechanism for a deep segmentation model, IEEe Access 10 (2022) 108205--108215.

\bibitem{62}
G.~Sun, Y.~Pan, W.~Kong, Z.~Xu, J.~Ma, T.~Racharak, L.-M. Nguyen, J.~Xin, Da-transunet: integrating spatial and channel dual attention with transformer u-net for medical image segmentation, Frontiers in Bioengineering and Biotechnology 12 (2024) 1398237.

\bibitem{63}
H.~Sun, J.~Xu, Y.~Duan, Paratranscnn: Parallelized transcnn encoder for medical image segmentation, arXiv preprint arXiv:2401.15307 (2024).

\bibitem{45}
Y.~Chang, H.~Menghan, Z.~Guangtao, Z.~Xiao-Ping, Transclaw u-net: Claw u-net with transformers for medical image segmentation, arXiv preprint arXiv:2107.05188 (2021).

\bibitem{64}
R.~Azad, M.~Heidari, M.~Shariatnia, E.~K. Aghdam, S.~Karimijafarbigloo, E.~Adeli, D.~Merhof, Transdeeplab: Convolution-free transformer-based deeplab v3+ for medical image segmentation, in: International Workshop on PRedictive Intelligence In MEdicine, Springer, 2022, pp. 91--102.

\bibitem{65}
R.~Azad, Y.~Jia, E.~K. Aghdam, J.~Cohen-Adad, D.~Merhof, Enhancing medical image segmentation with transception: A multi-scale feature fusion approach, arXiv preprint arXiv:2301.10847 (2023).

\bibitem{66}
X.~Huang, Z.~Deng, D.~Li, X.~Yuan, Y.~Fu, Missformer: An effective transformer for 2d medical image segmentation, IEEE Transactions on Medical Imaging 42~(5) (2022) 1484--1494.

\bibitem{23}
Y.~Wu, K.~Liao, J.~Chen, J.~Wang, D.~Z. Chen, H.~Gao, J.~Wu, D-former: A u-shaped dilated transformer for 3d medical image segmentation, Neural Computing and Applications 35~(2) (2023) 1931--1944.

\bibitem{25}
A.~Hatamizadeh, Z.~Xu, D.~Yang, W.~Li, H.~Roth, D.~Xu, Unetformer: A unified vision transformer model and pre-training framework for 3d medical image segmentation, arXiv preprint arXiv:2204.00631 (2022).

\bibitem{26}
Y.~Jiang, Y.~Zhang, X.~Lin, J.~Dong, T.~Cheng, J.~Liang, Swinbts: A method for 3d multimodal brain tumor segmentation using swin transformer, Brain sciences 12~(6) (2022) 797.

\bibitem{28}
L.~Xia, H.~Zhang, Y.~Wu, R.~Song, Y.~Ma, L.~Mou, J.~Liu, Y.~Xie, M.~Ma, Y.~Zhao, 3d vessel-like structure segmentation in medical images by an edge-reinforced network, Medical Image Analysis 82 (2022) 102581.

\bibitem{29}
A.~Shaker, M.~Maaz, H.~Rasheed, S.~Khan, M.-H. Yang, F.~S. Khan, Unetr++: delving into efficient and accurate 3d medical image segmentation, arXiv preprint arXiv:2212.04497 (2022).

\bibitem{67}
P.~Dong, H.~Niu, Z.~Yi, X.~Xu, nmpls-net: Segmenting pulmonary lobes using nmode, Mathematics 11~(22) (2023) 4675.

\bibitem{68}
S.~Wang, Y.~Chen, Z.~Yi, nmode-unet: A novel network for semantic segmentation of medical images, Applied Sciences 14~(1) (2024) 411.

\bibitem{69}
J.~Hu, C.~Yu, Z.~Yi, H.~Zhang, Enhancing robustness of medical image segmentation model with neural memory ordinary differential equation., International Journal of Neural Systems (2023) 2350060--2350060.

\bibitem{30}
H.~Pinckaers, G.~Litjens, Neural ordinary differential equations for semantic segmentation of individual colon glands, arXiv preprint arXiv:1910.10470 (2019).

\bibitem{31}
R.~Valle, F.~Reda, M.~Shoeybi, P.~Legresley, A.~Tao, B.~Catanzaro, Neural odes for image segmentation with level sets, arXiv preprint arXiv:1912.11683 (2019).

\bibitem{32}
R.~T. Chen, Y.~Rubanova, J.~Bettencourt, D.~K. Duvenaud, Neural ordinary differential equations, Advances in neural information processing systems 31 (2018).

\bibitem{33}
X.~Liu, T.~Xiao, S.~Si, Q.~Cao, S.~Kumar, C.-J. Hsieh, How does noise help robustness? explanation and exploration under the neural sde framework, in: Proceedings of the IEEE/CVF Conference on Computer Vision and Pattern Recognition, 2020, pp. 282--290.

\bibitem{34}
Q.~Kang, Y.~Song, Q.~Ding, W.~P. Tay, Stable neural ode with lyapunov-stable equilibrium points for defending against adversarial attacks, Advances in Neural Information Processing Systems 34 (2021) 14925--14937.

\bibitem{50}
W.~Wang, C.~Chen, M.~Ding, H.~Yu, S.~Zha, J.~Li, Transbts: Multimodal brain tumor segmentation using transformer, in: Medical Image Computing and Computer Assisted Intervention--MICCAI 2021: 24th International Conference, Strasbourg, France, September 27--October 1, 2021, Proceedings, Part I 24, Springer, 2021, pp. 109--119.

\bibitem{46}
G.~Xu, X.~Wu, X.~Zhang, X.~He, Levit-unet: Make faster encoders with transformer for medical image segmentation, arXiv preprint arXiv:2107.08623 (2021).

\bibitem{22}
H.-Y. Zhou, J.~Guo, Y.~Zhang, L.~Yu, L.~Wang, Y.~Yu, nnformer: Interleaved transformer for volumetric segmentation, arXiv preprint arXiv:2109.03201 (2021).

\bibitem{53}
Y.~Zhang, H.~Liu, Q.~Hu, Transfuse: Fusing transformers and cnns for medical image segmentation, in: Medical Image Computing and Computer Assisted Intervention--MICCAI 2021: 24th International Conference, Strasbourg, France, September 27--October 1, 2021, Proceedings, Part I 24, Springer, 2021, pp. 14--24.

\bibitem{54}
J.~Deng, W.~Dong, R.~Socher, L.-J. Li, K.~Li, L.~Fei-Fei, Imagenet: A large-scale hierarchical image database, in: 2009 IEEE conference on computer vision and pattern recognition, Ieee, 2009, pp. 248--255.

\bibitem{55}
L.~Bottou, Stochastic gradient descent tricks, in: Neural Networks: Tricks of the Trade: Second Edition, Springer, 2012, pp. 421--436.

\bibitem{38}
O.~Ronneberger, P.~Fischer, T.~Brox, U-net: Convolutional networks for biomedical image segmentation, in: Medical Image Computing and Computer-Assisted Intervention--MICCAI 2015: 18th International Conference, Munich, Germany, October 5-9, 2015, Proceedings, Part III 18, Springer, 2015, pp. 234--241.

\bibitem{39}
J.~Schlemper, O.~Oktay, M.~Schaap, M.~Heinrich, B.~Kainz, B.~Glocker, D.~Rueckert, Attention gated networks: Learning to leverage salient regions in medical images, Medical image analysis 53 (2019) 197--207.

\bibitem{43}
J.~Chen, Y.~Lu, Q.~Yu, X.~Luo, E.~Adeli, Y.~Wang, L.~Lu, A.~L. Yuille, Y.~Zhou, Transunet: Transformers make strong encoders for medical image segmentation, arXiv preprint arXiv:2102.04306 (2021).

\bibitem{14}
H.~Cao, Y.~Wang, J.~Chen, D.~Jiang, X.~Zhang, Q.~Tian, M.~Wang, Swin-unet: Unet-like pure transformer for medical image segmentation, in: European conference on computer vision, Springer, 2022, pp. 205--218.

\bibitem{48}
X.~Huang, Z.~Deng, D.~Li, X.~Yuan, Missformer: An effective medical image segmentation transformer, arXiv preprint arXiv:2109.07162 (2021).

\bibitem{36}
F.~Milletari, N.~Navab, S.-A. Ahmadi, V-net: Fully convolutional neural networks for volumetric medical image segmentation, in: 2016 fourth international conference on 3D vision (3DV), Ieee, 2016, pp. 565--571.

\bibitem{41}
O.~Oktay, J.~Schlemper, L.~L. Folgoc, M.~Lee, M.~Heinrich, K.~Misawa, K.~Mori, S.~McDonagh, N.~Y. Hammerla, B.~Kainz, et~al., Attention u-net: Learning where to look for the pancreas, arXiv preprint arXiv:1804.03999 (2018).

\end{thebibliography}
\end{document}